\documentclass{article} 
\usepackage[preprint]{colm2026_conference}
\usepackage{microtype}
\usepackage{hyperref}
\usepackage{url}
\usepackage{booktabs}
\usepackage{graphicx}
\usepackage{amsmath}
\usepackage{lineno}
\usepackage{tabularx}
\definecolor{darkblue}{rgb}{0, 0, 0.5}
\hypersetup{colorlinks=true, citecolor=darkblue, linkcolor=darkblue, urlcolor=darkblue}

\title{Domain-Adapted Retrieval for In-Context Annotation of Pedagogical Dialogue Acts}

\author{Jinsook Lee, Kirk Vanacore, Zhuqian Zhou, Bakhtawar Ahtisham \& Rene F. Kizilcec \\
Cornell University\\
Ithaca, NY, 14850, USA \\
\texttt{\{jl3369, kpv27, zz968, ba453, kizilcec\}@cornell.edu} \\
}

\begin{document}
\ifcolmsubmission
\linenumbers
\fi
\maketitle
\begin{abstract}
Automated annotation of pedagogical dialogue is a high-stakes task where LLMs often fail without sufficient domain grounding.We present a domain-adapted RAG pipeline for tutoring move annotation. Rather than fine-tuning the generative model, we adapt retrieval by fine-tuning a lightweight embedding model on tutoring corpora and indexing dialogues at the utterance level to retrieve labeled few-shot demonstrations. Evaluated across two real tutoring dialogue datasets (TalkMoves and Eedi) and three LLM backbones (GPT-5.2, Claude Sonnet 4.6, Qwen3-32b), our best configuration achieves Cohen's $\kappa$ of 0.526-0.580 on TalkMoves and 0.659-0.743 on Eedi, substantially outperforming no-retrieval baselines ($\kappa = 0.275$-$0.413$ and $0.160$-$0.410$). An ablation study reveals that utterance-level indexing, rather than embedding quality alone, is the primary driver of these gains, with top-1 label match rates improving from 39.7\% to 62.0\% on TalkMoves and 52.9\% to 73.1\% on Eedi under domain-adapted retrieval. Retrieval also corrects systematic label biases present in zero-shot prompting and yields the largest improvements for rare and context-dependent labels. These findings suggest that adapting the retrieval component alone is a practical and effective path toward expert-level pedagogical dialogue annotation while keeping the generative model frozen.
\end{abstract}

\section{Introduction}\label{sec:intro}
Large language models (LLMs) are increasingly deployed across educational technology for diverse applications, from generating instructional materials and supporting assessment design to real-time tutoring \citep{scarlatos2025exploring}. While prior work has examined how to adapt or optimize models for specific educational tasks, far less is known about how well LLMs perform when interpreting authentic pedagogical interactions \citep{chiang2023large, he2024annollm, anagnostopoulou2025human}. This gap is particularly important because educational applications often require understanding nuanced, context-dependent concepts that differ fundamentally from general-purpose language understanding tasks.
For example, tutoring moves capture the communicative intent behind tutor and student utterances in educational interactions. A tutor might scaffold a student's reasoning through guided questions, directly provide an answer, press for deeper justification, or manage the conversational flow. Studies demonstrate that certain tutoring moves, such as scaffolding, correlate with deeper conceptual understanding, while excessive telling does not \citep{michaels2008deliberative, o2015teacher, litman2005dialogue}. Therefore identifying these moves reliably is essential for evaluating tutoring quality at scale, providing formative feedback to human instructors, and training the reward models that support intelligent tutoring systems.
However, tutoring move annotation remains a manual, expert-intensive process. Codebooks such as the TalkMoves framework \citep{suresh2022talkmoves} require trained annotators who understand pedagogical theory, and achieving acceptable inter-rater reliability demands extensive calibration sessions \citep{vail2014refinement}. This bottleneck limits the scale at which tutoring interactions can be analyzed, a constraint that grows increasingly untenable as online tutoring platforms generate millions of dialogue sessions per year.
The challenge is that tutoring moves are abstract and inherently contextual. The distinction between those dialogue acts depends on surrounding conversational context, the student's prior responses, and the tutor's pedagogical intent rather than the surface form of a single utterance. While LLMs can classify general dialogue acts with reasonable accuracy \citep{shan2023annotating, tan2023informativeness}, general-purpose schemas do not capture the pedagogical subtlety required for educational dialogue.
We address these limitations with a three-stage approach that decouples domain adaptation from generation. In the first stage, we fine-tune a lightweight sentence embedding model on tutoring dialogue corpora with Multiple Negatives Ranking Loss (MNRL), adapting the representation space so that utterances serving similar pedagogical functions cluster together regardless of surface lexical variation, and use these embeddings to segment dialogues into coherent chunks for FAISS indexing. In the second stage, we retrieve the most semantically similar labeled examples from the index. In the third stage, retrieved examples are presented as few-shot demonstrations alongside the annotation codebook to a general-purpose LLM, which performs classification through in-context learning (ICL) \citep{lewis2020rag, chen2024retrievalstyleincontextlearningfewshot}. All domain adaptation resides in the retriever that makes the pipeline portable across LLM backbones and reusable as new models become available. Our contributions are as follows.
\begin{enumerate}
    \item A \textbf{three-stage retrieval-augmented annotation pipeline} that combines domain-adapted embedding, utterance-level indexing with parent chunk context, and codebook-grounded in-context learning with a frozen LLM. Our experiments show 81\% relative improvement in Cohen's $\kappa$ over zero-shot baselines across three LLM backbones.
    \item An \textbf{utterance-level indexing strategy} that preserves label-specific signal during search while returning parent chunks as few-shot demonstrations, paired with a dynamic semantic chunking algorithm calibrated against sparse ground truth boundaries.
    \item A comprehensive \textbf{ablation study} across retrieval conditions, retrieval depth, and LLM backbones on two tutoring dialogue datasets, demonstrating that domain adaptation on the retriever side alone accounts for the majority of the performance gain over codebook-only prompting.
\end{enumerate}
\section{Related Work}
\label{sec:related}
\subsection{Dialogue Act Annotation in Educational Settings}
Dialogue act annotation has a long history in educational research. The Accountable Talk framework \citep{michaels2008deliberative} established the theoretical foundation for analyzing classroom discourse through the lens of talk moves that promote productive discussion. \citet{suresh2022talkmoves} operationalized this framework computationally by releasing the TalkMoves dataset, which contains K-12 mathematics lesson transcripts annotated at the utterance level for teacher and student discursive moves.
Early computational approaches relied on supervised models trained on these labeled corpora \citep{jensen2021deep}, with subsequent work addressing class imbalance through AUC optimization \citep{yang2023robust} and integrating multiple annotation perspectives to capture the multi-functionality of tutoring utterances \citep{naim2025actionable}. A consistent finding is that inter-rater reliability remains a bottleneck, requiring extensive calibration even with refined codebooks \citep{vail2014refinement}.
\subsection{Annotating Dialogue Acts with LLMs}
Researchers increasingly rely on LLMs to scale annotation, though performance relative to human experts varies by task complexity and workflow design \citep{su2025large, yu2024assessing}. While LLMs perform well in constrained settings, complex discourse annotation typically requires structured pipelines to accommodate real-world taxonomies and multi-label cases \citep{ostyakova2023chatgpt}. A growing body of work frames LLM annotation as a \textit{process design} problem, improving robustness through ensembling and orchestration \citep{na2025llm, ahtisham2025aiannotationorchestrationevaluating}. Particularly in education, some approaches range from using GPT-family models as crowdsourced annotators \citep{he2024annollm} to generating synthetic tutoring dialogue data to improve coverage of rare codes \citep{shan2023annotating} and applying active learning to reduce annotation costs \citep{tan2023informativeness}. Frameworks like EduDCM further improve reliability by decomposing constructs into discrete components such as acts and events, enforcing consistency checks across multiple LLMs \citep{qi2024edudcm}.
Despite these advances, LLMs still struggle with multi-party and goal-oriented dialogue act classification where overlapping interactional threads and complex schemas exceed current capabilities \citep{qamar-etal-2025-llms, labruna2023unraveling}. Comparisons of human and LLM assessment find that models struggle with pedagogically nuanced distinctions without domain-specific guidance \citep{anagnostopoulou2025human}, motivating human-in-the-loop pipelines that integrate verification steps \citep{10.1145/3731120.3744574, shah2025using}.
\subsection{Retrieval-Based In-Context Learning and Domain Adaptation}
Retrieval-augmented generation (RAG), introduced by \citet{lewis2020rag}, grounds model outputs
in retrieved external evidence at inference time rather than relying solely on parametric knowledge.
This is especially consequential for annotation tasks involving rare or fine-grained categories,
where \citet{li2024longtail} showed selective retrieval yields the largest gains. \citet{chen2024retrievalstyleincontextlearningfewshot} extended this to classification via retrieval databases of labeled demonstrations, while \citet{agarwal2024many} and \citet{dong2024survey} both highlight that demonstration selection is among the most consequential design choices in ICL. How documents are segmented into retrieval units further shapes performance \citep{wang2025segmentation}, and in tutoring dialogue this is particularly high-stakes since fixed-size windows routinely fragment
coherent instructional exchanges.
The effectiveness of RAG also depends on the quality of the retrieval embedding space. General-purpose embeddings often fail to capture domain-specific semantic similarity, returning examples that are lexically similar but functionally irrelevant. Adapting the retriever to the
target domain through unsupervised LLM adaptation \citep{li2024llama2vec, behnamghader2024llm2vec}
or self-training \citep{xu2025simrag} is often more impactful than adapting the generator. 

\section{Method}
\label{sec:method}
\begin{figure}[htpb]
    \centering
    \includegraphics[width=1.0\textwidth, trim=0 35 0 35, clip]{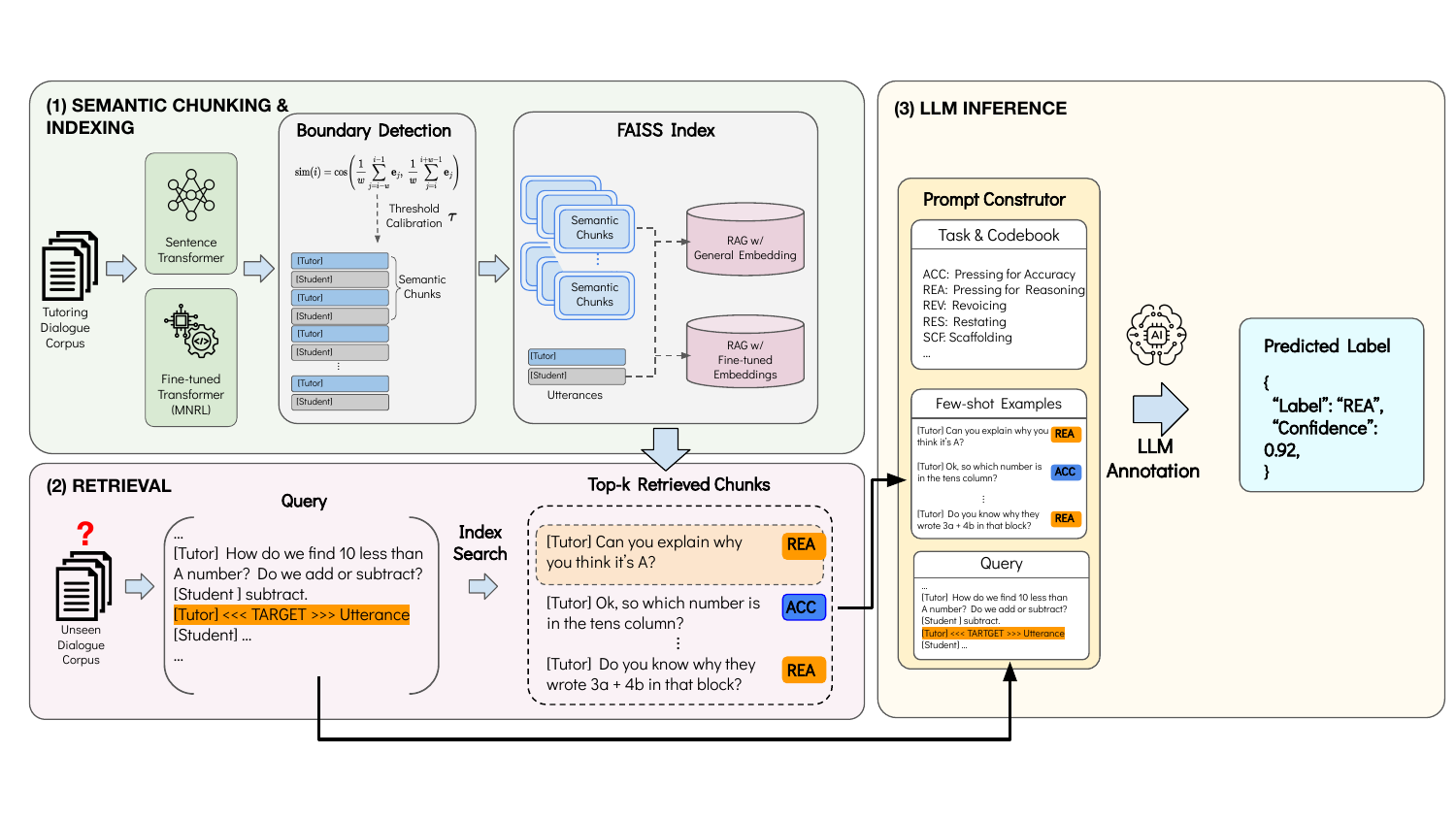}
    \caption{Overview of the proposed RAG-based annotation pipeline.}
    \label{fig:pdfimage}
\end{figure}
\subsection{Task Formulation}
Figure~\ref{fig:pdfimage} shows the overview of the proposed annotation pipeline. Each utterance in a tutoring dialogue has associated text $t_i$, a speaker role (tutor or student), and optionally a sparse ground-truth label from a predefined codebook. The label set follows the TalkMoves taxonomy with six categories \citep{suresh2022talkmoves}, namely Keeping Everyone Together (KET), Getting Students to Relate (GSR), Restating (RES), Pressing for Accuracy (ACC), Revoicing (REV), and Pressing for Reasoning (REA). The task is to predict the label for each tutor utterance by constructing a retrieval index from the corpus, dynamically building context windows around target utterances, and passing both local context and retrieved examples to a frozen LLM for in-context classification.
\subsection{Dynamic Semantic Chunking}
\label{sec:chunking}
When applying RAG, setting proper chunking boundaries is important \citep{wang2025segmentation, bhat2025rethinking}. If a chunk is too small, not enough information would be retrieved and if a chunk is too big, the information would be diluted. Our approach begins by partitioning the dialogue corpus into semantically coherent chunks that preserve label homogeneity while respecting session boundaries. We embed all utterances using a sentence transformer model $\phi$ such that $\mathbf{e}_i = \phi(t_i)$ where $\mathbf{e}_i \in \mathbb{R}^d$, L2-normalized to enable cosine similarity as a dot product. Semantic boundaries are detected by computing the smoothed cosine similarity between overlapping context windows. For each position $i$ in the dialogue, we compute
\begin{equation}
\text{sim}(i) = \cos\!\left(\frac{1}{w}\sum_{j=i-w}^{i-1}\mathbf{e}_j,\;\frac{1}{w}\sum_{j=i}^{i+w-1}\mathbf{e}_j\right)
\end{equation}
where $w$ is the window size (default $w = 2$). A low similarity score at position $i$ indicates a candidate semantic boundary where the topic or pedagogical function shifts.
To determine the boundary threshold $\tau$, we leverage sparse ground truth annotations. We identify true boundaries as positions where consecutive labeled utterances have differing labels, and non-boundaries as positions where consecutive labeled utterances share the same label. We then sweep $\tau$ over $[0.3, 1.0)$ in increments of $0.01$ and select the threshold that maximizes F1 on this boundary classification problem. We additionally impose a safety cap where $\tau$ cannot exceed the median of all observed similarity scores, preventing over-segmentation when fine-tuned embeddings exhibit representation collapse.
Chunks are created by splitting at positions where $\text{sim}(i) < \tau$, respecting a minimum size of 2 and maximum of 20 utterances.
\subsection{Domain-Adapted Embeddings}
\label{sec:embeddings}
To improve semantic separation of tutoring dialogue labels in the embedding space, we fine-tune \texttt{BGE-large-en-v1.5} on labeled utterances from the TalkMoves and Eedi training sets combined. We combine both datasets because they share the same label taxonomy, and pooling increases the diversity of in-batch negatives for MNRL while improving coverage of rare labels that are underrepresented in either dataset alone. Training on both classroom and dyadic chat data also encourages the embedding space to capture pedagogical function across interaction formats rather than surface patterns specific to one setting. To prevent data leakage through the embedding geometry, we use only training-split utterances for fine-tuning.
Training data consists of (anchor, positive) pairs sampled within each label group, capped at 3,000 pairs per label to prevent frequent labels from dominating the gradient. We optimize with MNRL \citep{henderson2017efficient}, where all other in-batch examples serve as implicit negatives. For a batch of $B$ pairs, the loss for pair $i$ is
\begin{equation}
\mathcal{L}_i = -\log \frac{\exp(\text{sim}(\mathbf{a}_i, \mathbf{p}_i) / T)}{\sum_{j=1}^{B} \exp(\text{sim}(\mathbf{a}_i, \mathbf{p}_j) / T)}
\end{equation}
where $\text{sim}$ is cosine similarity and $T$ is a temperature parameter. Training proceeds for 3 epochs with batch size 64 and we saved the best checkpoint by evaluation Spearman correlation.
\subsection{RAG Index Construction}
\label{sec:index}
We construct two types of FAISS indexes \citep{johnson2021faiss} depending on the embedding condition. We construct three indexes. A chunk-level index using Cohere embeddings, and two BGE-based indexes. The first BGE index operates at the chunk level; the second embeds each labeled utterance individually, retrieving the parent chunk at query time. At query time, we retrieve the top $k$ most similar utterances and return their parent chunks as few-shot demonstrations.
MNRL optimizes same-label similarity but does not explicitly enforce separation between different labels, which can cause representation collapse where all embeddings cluster into a narrow cone with pathologically high pairwise similarity. To address this, we apply mean-centering by subtracting the mean embedding vector from all stored embeddings before indexing. The same mean vector is subtracted from query embeddings at search time.

\subsection{Query Construction and Prompting}
\label{sec:query}
At inference time, we construct a context window for each target utterance that adapts to the dialogue structure. Given a target utterance at position $t$, we expand backward and forward from $t$, stopping at positions where $\text{sim}(i) < \tau$ or at session boundaries, up to a maximum of 10 utterances in each direction. For chunk-level indexes, we embed the full context window as the query vector. For utterance-level indexes, we embed only the target utterance itself.
Given the query, we search the FAISS index for the top $k$ entries (default $k = 3$). For utterance-level indexes, each retrieved utterance's parent chunk is returned as the demonstration, providing the LLM with full surrounding dialogue context. To avoid data leakage, we exclude the target utterance's own session from retrieval results.
The LLM prompt consists of a system message specifying the annotator role, task definition, and full codebook with label definitions, followed by a user message presenting retrieved chunks as labeled examples and the current dialogue context with the target utterance marked by \texttt{<<<TARGET>>>}. The model outputs a JSON object with the predicted label and confidence score. Full prompt templates are provided in Appendix~\ref{app:prompts}.
\section{Experimental Setup}
\label{sec:experiments}
\subsection{Datasets}
\paragraph{TalkMoves.} The TalkMoves dataset\footnote{\url{https://github.com/SumnerLab/TalkMoves/tree/main}} contains human-transcribed K-12 mathematics classroom transcripts drawn from whole-class discussions, small-group problem-solving, and online lesson contexts \citep{suresh2022talkmoves}. Utterances are annotated with six talk moves organized into three dimensions: \emph{Learning Community} (KET, GSR, RES), \emph{Content Knowledge} (ACC), and \emph{Rigorous Thinking} (REV, REA). We focus exclusively on teacher talk moves. After removing 2 sessions with fewer than 10 human labels, we retain 61 sessions split into 49 training and 12 test (Table~\ref{tab:datasets}). The label distribution is heavily imbalanced, with ACC (41\%) and KET (27\%) dominating while REA and GSR each comprise fewer than 6\%.
\paragraph{Eedi.} The Eedi dataset \citep{zent2025eedi}\footnote{\url{https://huggingface.co/datasets/Eedi/Question-Anchored-Tutoring-Dialogues-2k}} is drawn from a UK-based online tutoring platform deployed in over 19,000 schools worldwide. Conversations are initiated when a student requests help from an on-demand expert tutor while working on a math lesson, producing dyadic chat-based dialogues anchored to a specific question. The data were collected from November 2021 to February 2025 and annotated with the same six TalkMoves categories used for annotation.
\begin{table}[htpb!]
\centering
\small
\begin{tabular}{lrrrrrrr}
\toprule
Dataset & Sessions & Utterances & Avg. Utt & Annotated (\%) & Train & Test \\
\midrule
TalkMoves & 61 & 31,171 & 511 & 8,407 (27\%) & 49 & 12 \\
Eedi & 1,965 & 46,147 & 23 & 30,001 (65\%) & 1,570 & 395 \\
\bottomrule
\end{tabular}
\caption{Dataset statistics. Annotated counts and percentages reflect ground-truth labels across the full dataset (train and test combined). Avg. Utt is the mean utterances per session.}
\label{tab:datasets}
\end{table}
\subsection{Conditions and Implementation}
We compare four retrieval conditions (Table~\ref{tab:conditions}). The \textsc{no\_rag} condition provides only the codebook and local dialogue context. The \textsc{RAG\_no\_finetune} condition uses Cohere \texttt{embed-english-v3} for semantic chunking and chunk-level retrieval. The \textsc{RAG\_finetuned\_chunk} condition replaces the embedding model with our fine-tuned BGE while retaining chunk-level indexing, isolating the effect of embedding quality from index granularity. The \textsc{RAG\_finetuned\_utt} condition uses the same fine-tuned BGE model but indexes at the utterance level with parent chunk retrieval. We additionally evaluate fixed-window chunking baselines (Appendix~\ref{app:fixedwindow}).
\begin{table}[htpb!]
\centering
\small
\begin{tabularx}{\columnwidth}{lllll}
\toprule
Condition & Chunking & Retrieval & Index & Returned \\
\midrule
\textsc{no\_rag} & None & None (codebook only) & -- & -- \\
\textsc{RAG\_no\_finetune} & Semantic (Cohere) & Cohere embed-english-v3 & Chunk & Chunk \\
\textsc{RAG\_finetuned\_chunk} & Semantic (FT-BGE) & Fine-tuned BGE (MNRL) & Chunk & Chunk \\
\textsc{RAG\_finetuned\_utt} & Semantic (FT-BGE) & Fine-tuned BGE (MNRL) & Utt. & Chunk \\
\bottomrule
\end{tabularx}
\caption{Experimental conditions. Index denotes the granularity of FAISS entries. Returned denotes what is passed to the LLM as a few-shot demonstration. \textsc{RAG\_finetuned\_chunk} and \textsc{RAG\_finetuned\_utt} share the same embedding model, isolating the effect of index granularity. Additional ablation conditions are reported in Appendix~\ref{app:fixedwindow}.}
\label{tab:conditions}
\end{table}
To test the generality of the pipeline across LLM backbones, we evaluate the three main conditions with three models: GPT-5.2, Claude Sonnet 4.6, and Qwen3-32b. All models are accessed via LiteLLM's unified interface with temperature 1.0, which we retain as the default to avoid artificially constraining the model's output distribution. Lower temperatures may improve consistency but risk suppressing valid alternative predictions for ambiguous utterances. The default retrieval depth is $k = 3$, with ablation over $k \in \{0, 1, 3, 5\}$ for all three models. Our primary metric is Cohen's $\kappa$, computed only on utterances with human labels, excluding null predictions. We additionally report per-label precision, recall, and F1. Statistical significance is assessed via permutation tests ($n = 2{,}000$) for $\kappa$ and McNemar's test for accuracy. For embedding fine-tuning, we use \texttt{BGE-large-en-v1.5} \citep{xiao2023cpack} with batch size 64 for 3 epochs.

\section{Results}
\label{sec:results}

\subsection{Domain-Adapted Retrieval Consistently Improves Annotation Quality Across All Models and Datasets}

\begin{table}[htpb!]
\centering
\small
\begin{tabular}{llcccc}
\toprule
Model & Condition & TM $\kappa$ & TM Acc & Eedi $\kappa$ & Eedi Acc \\
\midrule
GPT-5.2 & \textsc{no\_rag} & 0.315 & 0.500 & 0.351 & 0.575 \\
GPT-5.2 & \textsc{RAG\_no\_finetune} & 0.479 & 0.639 & 0.505 & 0.709 \\
GPT-5.2 & \textsc{RAG\_finetuned\_chunk} & 0.454 & 0.620 & 0.530 & 0.740 \\
GPT-5.2 & \textsc{RAG\_finetuned\_utt} & 0.539 & 0.688 & 0.659 & 0.810 \\
\midrule
Sonnet 4.6 & \textsc{no\_rag} & 0.413 & 0.578 & 0.410 & 0.714 \\
Sonnet 4.6 & \textsc{RAG\_no\_finetune} & 0.498 & 0.653 & 0.632 & 0.797 \\
Sonnet 4.6 & \textsc{RAG\_finetuned\_chunk} & 0.501 & 0.647 & 0.632 & 0.780 \\
Sonnet 4.6 & \textsc{RAG\_finetuned\_utt} & \textbf{0.580} & \textbf{0.708} & \textbf{0.743} & \textbf{0.847} \\
\midrule
Qwen3-32b & \textsc{no\_rag} & 0.275 & 0.404 & 0.160 & 0.354 \\
Qwen3-32b & \textsc{RAG\_no\_finetune} & 0.442 & 0.617 & 0.590 & 0.769 \\
Qwen3-32b & \textsc{RAG\_finetuned\_chunk} & 0.428 & 0.585 & 0.545 & 0.725 \\
Qwen3-32b & \textsc{RAG\_finetuned\_utt} & 0.526 & 0.667 & 0.694 & 0.818 \\
\bottomrule
\end{tabular}
\caption{Cohen's $\kappa$ and accuracy across three LLM backbones on both test sets ($k = 3$). Bold indicates the best result per metric. All RAG vs. \textsc{no\_rag} differences are significant at $p < .001$ (permutation test, $n = 2{,}000$) except Qwen3-32b $\kappa$ on TalkMoves for \textsc{RAG\_no\_finetune} ($p < .01$).}
\label{tab:main_results}
\end{table}

Table~\ref{tab:main_results} presents Cohen's $\kappa$ and accuracy across all three models and 
retrieval conditions on both test sets with $k = 3$. Across all models, retrieval-augmented 
prompting substantially improves annotation quality over the codebook-only baseline, and 
domain-adapted retrieval (\textsc{RAG\_finetuned\_utt}) consistently outperforms general-purpose 
retrieval (\textsc{RAG\_no\_finetune}). This pattern holds regardless of the underlying LLM's 
baseline capability: Qwen3-32b, despite the weakest no-retrieval performance ($\kappa = 0.275$ on 
TalkMoves, $0.160$ on Eedi), reaches $\kappa = 0.526$ and $0.694$ with domain-adapted retrieval, 
competitive with GPT-5.2's retrieval-augmented results. Claude Sonnet 4.6 achieves the highest 
overall performance at $\kappa = 0.580$ on TalkMoves and $0.743$ on Eedi. The gains on Eedi are 
notably larger across all models, likely because the one-on-one tutoring format produces more 
focused exchanges where utterance-level retrieval can find closer semantic matches.

\subsection{Retrieval Precision Is the Mechanistic Driver of Annotation Gains}

\begin{table}[htbp!]
\centering
\small
\begin{tabular}{lcccc}
\toprule
 & \multicolumn{2}{c}{TalkMoves} & \multicolumn{2}{c}{Eedi} \\
\cmidrule(lr){2-3} \cmidrule(lr){4-5}
Condition & Top-1 & Any-$k$ & Top-1 & Any-$k$ \\
\midrule
\textsc{RAG\_no\_finetune} & 39.7\% & 63.4\% & 52.9\% & 75.6\% \\
\textsc{RAG\_finetuned\_utt} & 62.0\% & 83.4\% & 73.1\% & 89.9\% \\
\midrule
$\Delta$ & +22.3pp & +20.0pp & +20.2pp & +14.3pp \\
\bottomrule
\end{tabular}
\caption{Retrieval label match rate. Top-1 measures whether the nearest retrieved example shares the target's gold label. Any-$k$ ($k = 3$) measures whether at least one of the top-$k$ retrieved examples matches. Domain-adapted embeddings produce substantially more label-relevant demonstrations.}
\label{tab:retrieval_quality}
\end{table}

To verify that annotation gains stem from improved retrieval quality, Table~\ref{tab:retrieval_quality} 
reports the label match rate between retrieved demonstrations and the target utterance's gold label. 
With \textsc{RAG\_finetuned\_utt}, the top-1 match rate reaches 62.0\% on TalkMoves and 73.1\% on 
Eedi, compared to 39.7\% and 52.9\% for \textsc{RAG\_no\_finetune}. The effect is most pronounced 
for rare labels: on TalkMoves, the match rate for ACC jumps from 29\% to 79\%. Conversely, labels 
where retrieval quality remains low (e.g., RES on Eedi at 0\% for both conditions) correspond 
directly to near-zero annotation $\kappa$, confirming that retrieval precision drives annotation 
quality. Detailed per-label analysis is in Appendix~\ref{app:retrieval_quality}.

\subsection{Utterance-Level Indexing, Not Embedding Quality, Drives Performance Gains}

Because \textsc{RAG\_no\_finetune} and \textsc{RAG\_finetuned\_utt} differ in both embedding model 
and index granularity, we introduce \textsc{RAG\_finetuned\_chunk} to isolate each factor. Using 
the same fine-tuned BGE embeddings but chunk-level indexing, \textsc{RAG\_finetuned\_chunk} achieves 
$\kappa = 0.454$ on TalkMoves and $0.530$ on Eedi with GPT-5.2, nearly identical to 
\textsc{RAG\_no\_finetune} ($0.479$ and $0.505$). The large jump to \textsc{RAG\_finetuned\_utt} 
($0.539$ and $0.659$) therefore comes predominantly from utterance-level indexing ($+0.085$ and 
$+0.129$ respectively), with embedding quality contributing negligibly ($\Delta\kappa \approx 0$). 
This pattern holds across all three LLM backbones (Appendix~\ref{app:confound}), and performance 
improves monotonically with retrieval depth through $k = 5$, with the gap between conditions 
widening as $k$ increases (Appendix~\ref{app:k_ablation}).

\subsection{Retrieval Improves Rare and 
Context-Dependent Annotation Performance}

\begin{figure}[htpb!]
    \centering
    \includegraphics[width=\textwidth]{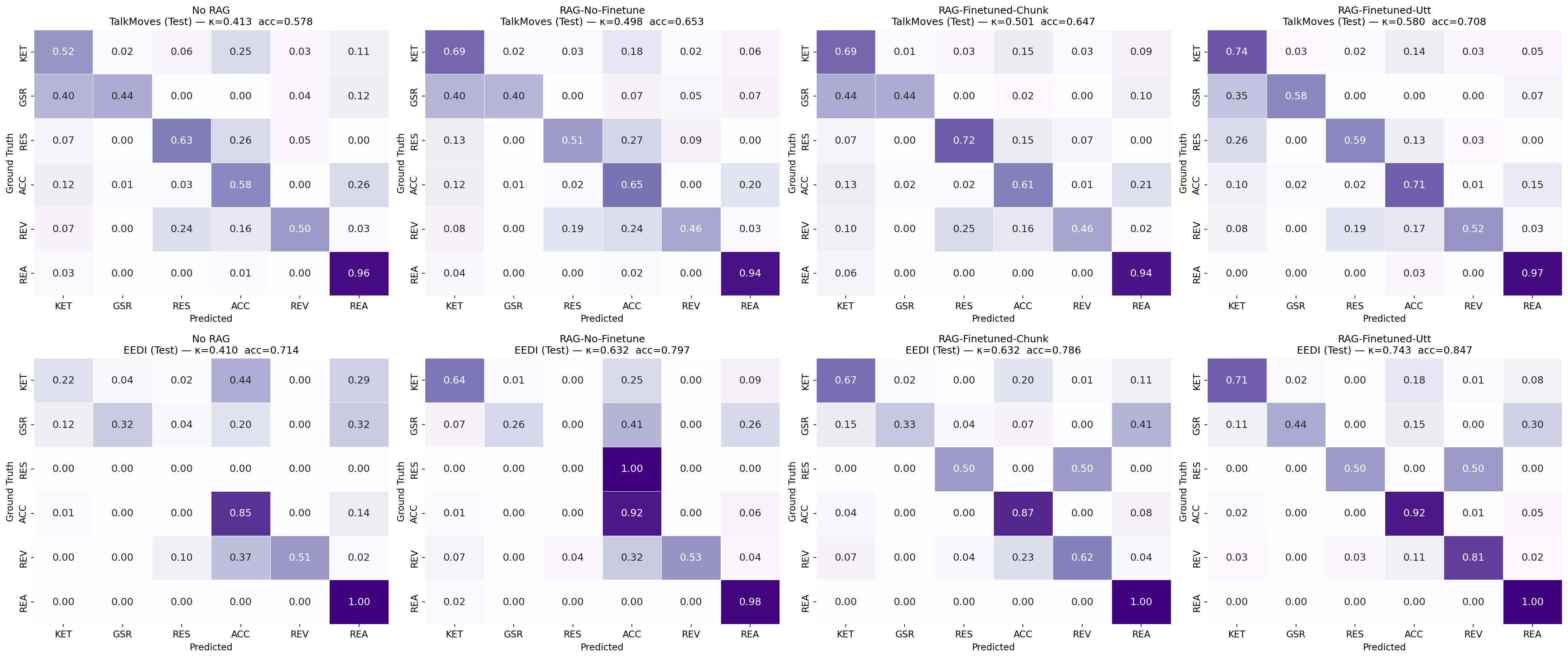}
    \caption{Normalized confusion matrices for TalkMoves test (top) and Eedi test (bottom) with Claude Sonnet 4.6. Domain-adapted retrieval (\textsc{RAG\_finetuned\_utt}, right) concentrates mass along the diagonal for all labels. Without retrieval, REA is systematically over-predicted on Eedi.}
    \label{fig:confusion}
\end{figure}

Figure~\ref{fig:confusion} presents normalized confusion matrices for Claude Sonnet 4.6, the 
best-performing backbone. Without retrieval, the model exhibits a strong bias toward REA on Eedi, 
assigning it to a substantial fraction of utterances from nearly every other category (e.g., 0.32 
of GSR and 0.14 of ACC), while on TalkMoves it over-predicts ACC and shows scattered 
misclassifications across rare labels. Retrieval largely eliminates these biases, with 
\textsc{RAG\_finetuned\_utt} strengthening diagonal entries across all labels, most notably for REV 
on Eedi (0.51 to 0.81) and KET on TalkMoves (0.52 to 0.74). The largest absolute gains appear for 
rare and context-dependent labels: ACC improves from 0.85 to 0.92 and GSR jumps from 0.32 to 0.44 
on Eedi, while GSR rises from 0.44 to 0.58 on TalkMoves. These patterns suggest that retrieval 
helps by providing disambiguating context rather than merely compensating for label scarcity.

Despite these gains, the most persistent confusion on TalkMoves is ACC predicted when the ground truth is KET (0.14 of KET utterances misclassified even with \textsc{RAG\_finetuned\_utt}), occurring when organizational utterances include question-like forms that superficially resemble 
accuracy checks. Rare labels also remain challenging, with REA accounting for fewer than 5\% of 
annotations in both datasets. Per-label $\kappa$ values and full classification reports are in 
Appendix~\ref{app:results}.

\section{Discussion}
\label{sec:discussion}
\paragraph{Utterance-level indexing as the primary lever for improvement.}
Our confound isolation analysis reveals that the dominant performance gain comes from utterance-level indexing rather than embedding quality. When the fine-tuned BGE model is used with chunk-level indexing (\textsc{RAG\_finetuned\_chunk}), $\kappa$ is nearly identical to \textsc{RAG\_no\_finetune} ($\Delta\kappa \approx 0$ on both datasets). The large jump to \textsc{RAG\_finetuned\_utt} ($+0.085$ on TalkMoves, $+0.129$ on Eedi for GPT-5.2) comes from switching the index granularity from chunks to individual utterances. Chunk-level embeddings average over multiple utterances and dialogue acts, diluting the label-specific signal that drives accurate retrieval. Utterance-level indexing preserves this signal, enabling the retriever to find demonstrations that share the target's pedagogical function rather than its surface topic. This finding aligns with the broader principle that adapting the retriever is more impactful than adapting the generator \citep{xu2025simrag, li2024llama2vec}, and refines it by showing that the indexing granularity matters more than the embedding model itself.
\paragraph{Robustness across LLM backbones.}
The consistent improvement pattern across GPT-5.2, Claude Sonnet 4.6, and Qwen3-32b confirms that our pipeline's gains are not an artifact of a particular model's strengths. Notably, Qwen3-32b has the weakest no-retrieval baseline but benefits the most from domain-adapted retrieval in relative terms, with $\kappa$ improving by over 4$\times$ on Eedi (from 0.160 to 0.694). This suggests that the pipeline is particularly valuable for models with weaker zero-shot classification capabilities, effectively compensating for limited parametric knowledge through high-quality retrieved demonstrations. Since all domain adaptation resides in the 335M-parameter embedding model and FAISS index, the pipeline can be paired with any LLM backend without retraining.
\paragraph{Chunking strategy is secondary to index granularity.}
The fixed-window ablation (Appendix~\ref{app:fixedwindow}) shows that the chunking algorithm itself has limited impact on downstream annotation quality. When the same retrieval model is used, fixed windows of $w = 3$ match or slightly outperform semantic chunking. This further supports the finding that index granularity is the primary design choice. Returning the parent chunk at inference time still provides the LLM with surrounding conversational context, bridging retrieval precision and contextual richness.
\paragraph{Limitations.}
Our evaluation is restricted to two mathematics tutoring datasets annotated with the same taxonomy, though TalkMoves captures multi-party classroom discussions while Eedi consists of dyadic online chat, and some categories (e.g., Getting Students to Relate) may function differently across these settings. Future work should evaluate generalization across different annotation taxonomies (e.g., STEM tutoring acts, Socratic dialogue codes) and subject domains beyond mathematics to determine whether domain-adapted retrieval transfers or requires taxonomy-specific re-training. We note that sample sizes vary across models and conditions due to differences in valid prediction rates. Because each condition is evaluated only on the subset of utterances for which it produced a valid prediction, conditions that fail on harder examples may report inflated metrics on the remaining easier subset.
\section{Conclusion}
\label{sec:conclusion}
We presented a retrieval-augmented pipeline for annotating pedagogical dialogue acts at scale. By combining domain-adapted embeddings, dynamic semantic chunking, and codebook-grounded prompting, our system achieves Cohen's $\kappa$ of up to 0.580 on TalkMoves and 0.743 on Eedi across three LLM backbones, substantially outperforming codebook-only baselines without task-specific LLM fine-tuning. Domain adaptation on the retriever side accounts for the majority of this gain, and the improvement pattern holds across models of varying capability. Utterance-level indexing with parent chunk retrieval consistently outperforms chunk-level indexing by preserving label-specific signal while providing rich conversational context. Future directions include extending to non-mathematical tutoring domains, integrating active learning for iterative index improvement, multi-turn annotation that captures dialogue-level pedagogical strategies, and alternative training objectives that enforce inter-label separation without post-hoc corrections.
\section{Ethics Statement}
We analyze de-identified tutoring dialogue datasets in accordance with our Institutional Review Board (IRB)-approved protocol [BLIND FOR REVIEW] . The TalkMoves dataset was de-identified by humans before it was open sourced, and the Eedi dataset was de-identified by the tutoring provider with participant consent for research use. We follow IRB-approved procedures for data storage and access and use these data solely to understand and improve dialogue annotation workflows. Re-identification attempts, user profiling, or any use that could enable harm to individual participants is prohibited. Although the data are de-identified, conversational text can carry residual privacy risk. We therefore minimize verbatim excerpts, avoid reporting sensitive attributes, and present results in aggregate wherever possible.
Automated annotation of tutoring moves is intended to \emph{augment} rather than replace expert human judgment. The confidence scores produced by our pipeline support a human-in-the-loop workflow in which uncertain predictions are flagged for manual review. We caution against deploying automated labels as ground truth without such verification, particularly for high-stakes decisions such as teacher evaluation or student assessment. Our system relies on commercial and open-weight LLMs accessed via API, which introduces dependence on external services and associated cost, latency, and data-handling considerations. Researchers applying this pipeline should ensure that any data sent to external APIs complies with applicable data governance policies. Finally, the label taxonomy was developed for mathematics tutoring in English and may not transfer to other languages, subject areas, or cultural contexts without careful adaptation and validation. We used Claude for editorial assistance in drafting and revising portions of this manuscript.
\bibliography{colm2026_conference}

\begin{thebibliography}{38}
\providecommand{\natexlab}[1]{#1}
\providecommand{\url}[1]{\texttt{#1}}
\expandafter\ifx\csname urlstyle\endcsname\relax
  \providecommand{\doi}[1]{doi: #1}\else
  \providecommand{\doi}{doi: \begingroup \urlstyle{rm}\Url}\fi

\bibitem[Agarwal et~al.(2024)Agarwal, Singh, Zhang, Bohnet, Rosias, Chan, Zhang, Anand, Abbas, Nova, Co-Reyes, Chu, Behbahani, Faust, and Larochelle]{agarwal2024many}
Rishabh Agarwal, Avi Singh, Lei Zhang, Bernd Bohnet, Luis Rosias, Stephanie Chan, Biao Zhang, Ankesh Anand, Zaheer Abbas, Azade Nova, John~D. Co-Reyes, Eric Chu, Feryal Behbahani, Aleksandra Faust, and Hugo Larochelle.
\newblock Many-shot in-context learning.
\newblock In \emph{Advances in Neural Information Processing Systems}, 2024.

\bibitem[Ahtisham et~al.(2025)Ahtisham, Vanacore, Lee, Zhou, Pietrzak, and Kizilcec]{ahtisham2025aiannotationorchestrationevaluating}
Bakhtawar Ahtisham, Kirk Vanacore, Jinsook Lee, Zhuqian Zhou, Doug Pietrzak, and Rene~F. Kizilcec.
\newblock Ai annotation orchestration: Evaluating llm verifiers to improve the quality of llm annotations in learning analytics, 2025.
\newblock URL \url{https://arxiv.org/abs/2511.09785}.

\bibitem[Anagnostopoulou et~al.(2025)Anagnostopoulou, Feldhus, Hsu, Alshomary, Wachsmuth, and Sonntag]{anagnostopoulou2025human}
Aliki Anagnostopoulou, Nils Feldhus, Yi-Sheng Hsu, Milad Alshomary, Henning Wachsmuth, and Daniel Sonntag.
\newblock Human and llm-based assessment of teaching acts in expert-led explanatory dialogues.
\newblock In \emph{Proceedings of the 6th Workshop on Computational Approaches to Discourse}, pp.\  166--181, 2025.

\bibitem[BehnamGhader et~al.(2024)BehnamGhader, Adlakha, Mosbach, Bahdanau, Chapados, and Reddy]{behnamghader2024llm2vec}
Parishad BehnamGhader, Vaibhav Adlakha, Marius Mosbach, Dzmitry Bahdanau, Nicolas Chapados, and Siva Reddy.
\newblock Llm2vec: Large language models are secretly powerful text encoders.
\newblock In \emph{Proceedings of the 1st Conference on Language Modeling}, 2024.

\bibitem[Bhat et~al.(2025)Bhat, Rudat, Spiekermann, and Flores-Herr]{bhat2025rethinking}
Sinchana~Ramakanth Bhat, Max Rudat, Jannis Spiekermann, and Nicolas Flores-Herr.
\newblock Rethinking chunk size for long-document retrieval: A multi-dataset analysis.
\newblock \emph{arXiv preprint arXiv:2505.21700}, 2025.

\bibitem[Chen et~al.(2024)Chen, Zhao, Chen, Wang, Li, Zhang, and Zhang]{chen2024retrievalstyleincontextlearningfewshot}
Huiyao Chen, Yu~Zhao, Zulong Chen, Mengjia Wang, Liangyue Li, Meishan Zhang, and Min Zhang.
\newblock Retrieval-style in-context learning for few-shot hierarchical text classification, 2024.
\newblock URL \url{https://arxiv.org/abs/2406.17534}.

\bibitem[Chiang \& Lee(2023)Chiang and Lee]{chiang2023large}
Cheng-Han Chiang and Hung-yi Lee.
\newblock Can large language models be an alternative to human evaluations?
\newblock In \emph{Proceedings of the 61st Annual Meeting of the Association for Computational Linguistics (Volume 1: Long Papers)}, pp.\  15607--15631, Toronto, Canada, July 2023. Association for Computational Linguistics.
\newblock \doi{10.18653/v1/2023.acl-long.870}.
\newblock URL \url{https://aclanthology.org/2023.acl-long.870/}.

\bibitem[Dong et~al.(2024)Dong, Li, Dai, Zheng, Ma, Li, Xia, Xu, Wu, Chang, Sun, Li, and Sui]{dong2024survey}
Qingxiu Dong, Lei Li, Damai Dai, Ce~Zheng, Jingyuan Ma, Rui Li, Heming Xia, Jingjing Xu, Zhiyong Wu, Baobao Chang, Xu~Sun, Lei Li, and Zhifang Sui.
\newblock A survey on in-context learning.
\newblock In \emph{Proceedings of the 2024 Conference on Empirical Methods in Natural Language Processing}, 2024.

\bibitem[He et~al.(2024)He, Lin, Gong, Jin, Zhang, Lin, Jiao, Yiu, Duan, and Chen]{he2024annollm}
Xingwei He, Zhenghao Lin, Yeyun Gong, A-Long Jin, Hang Zhang, Chen Lin, Jian Jiao, Siu~Ming Yiu, Nan Duan, and Weizhu Chen.
\newblock Annollm: Making large language models to be better crowdsourced annotators.
\newblock In \emph{Proceedings of the 2024 Conference of the North American Chapter of the Association for Computational Linguistics: Human Language Technologies (Volume 6: Industry Track)}, pp.\  165--190, Mexico City, Mexico, June 2024. Association for Computational Linguistics.
\newblock URL \url{https://aclanthology.org/2024.naacl-industry.15/}.

\bibitem[Henderson et~al.(2017)Henderson, Al-Rfou, Strope, Sung, Luk{\'a}cs, Guo, Kumar, Miklos, and Kurzweil]{henderson2017efficient}
Matthew Henderson, Rami Al-Rfou, Brian Strope, Yun-Hsuan Sung, L{\'a}szl{\'o} Luk{\'a}cs, Ruiqi Guo, Sanjiv Kumar, Balint Miklos, and Ray Kurzweil.
\newblock Efficient natural language response suggestion for smart reply.
\newblock 2017.
\newblock URL \url{https://arxiv.org/abs/1705.00652}.

\bibitem[Jensen et~al.(2021)Jensen, Pugh, and D'Mello]{jensen2021deep}
Emily Jensen, Samuel~L. Pugh, and Sidney~K. D'Mello.
\newblock A deep transfer learning approach to modeling teacher discourse in the classroom.
\newblock In \emph{Proceedings of the 11th International Learning Analytics and Knowledge Conference}, pp.\  302--312, 2021.

\bibitem[Johnson et~al.(2021)Johnson, Douze, and J{\'e}gou]{johnson2021faiss}
Jeff Johnson, Matthijs Douze, and Herv{\'e} J{\'e}gou.
\newblock Billion-scale similarity search with {GPUs}.
\newblock \emph{IEEE Transactions on Big Data}, 7\penalty0 (3):\penalty0 535--547, 2021.

\bibitem[Labruna et~al.(2023)Labruna, Brenna, Zaninello, and Magnini]{labruna2023unraveling}
Tiziano Labruna, Sofia Brenna, Andrea Zaninello, and Bernardo Magnini.
\newblock Unraveling chatgpt: A critical analysis of ai-generated goal-oriented dialogues and annotations.
\newblock In \emph{International Conference of the Italian Association for Artificial Intelligence}, pp.\  151--171. Springer, 2023.

\bibitem[Lewis et~al.(2020)Lewis, Perez, Piktus, Petroni, Karpukhin, Goyal, K\"{u}ttler, Lewis, Yih, Rockt\"{a}schel, Riedel, and Kiela]{lewis2020rag}
Patrick Lewis, Ethan Perez, Aleksandra Piktus, Fabio Petroni, Vladimir Karpukhin, Naman Goyal, Heinrich K\"{u}ttler, Mike Lewis, Wen-tau Yih, Tim Rockt\"{a}schel, Sebastian Riedel, and Douwe Kiela.
\newblock Retrieval-augmented generation for knowledge-intensive nlp tasks.
\newblock In \emph{Advances in Neural Information Processing Systems}, volume~33, pp.\  9459--9474, 2020.

\bibitem[Li et~al.(2024{\natexlab{a}})Li, Liu, Xiao, Shao, and Lian]{li2024llama2vec}
Chaofan Li, Zheng Liu, Shitao Xiao, Yingxia Shao, and Defu Lian.
\newblock Llama2vec: Unsupervised adaptation of large language models for dense retrieval.
\newblock In \emph{Proceedings of the 62nd Annual Meeting of the Association for Computational Linguistics}, 2024{\natexlab{a}}.

\bibitem[Li et~al.(2024{\natexlab{b}})Li, Yan, Zhang, Wang, He, Huang, Xue, and Huang]{li2024longtail}
Dongyang Li, Junbing Yan, Taolin Zhang, Cheng Wang, Xiaofeng He, Longtao Huang, Hui Xue, and Jun Huang.
\newblock On the role of long-tail knowledge in retrieval augmented large language models.
\newblock In \emph{Proceedings of the 62nd Annual Meeting of the Association for Computational Linguistics (Short Papers)}, 2024{\natexlab{b}}.

\bibitem[Litman \& Forbes-Riley(2005)Litman and Forbes-Riley]{litman2005dialogue}
Diane Litman and Kate Forbes-Riley.
\newblock Correlating student acoustic-prosodic profiles with student learning in spoken tutoring dialogues.
\newblock In \emph{Proceedings of the 12th International Conference on Artificial Intelligence in Education}, pp.\  1--8, 2005.

\bibitem[Michaels \& O'Connor(2015)Michaels and O'Connor]{o2015teacher}
S.~Michaels and C.~O'Connor.
\newblock Conceptualizing talk moves as tools: Professional development approaches for academically productive discussions.
\newblock In Lauren~B. Resnick, Christa Asterhan, and Sherice~N. Clarke (eds.), \emph{Socializing Intelligence through Talk and Dialogue}, pp.\  333--347. American Educational Research Association, Washington DC, 2015.

\bibitem[Michaels et~al.(2008)Michaels, O'Connor, and Resnick]{michaels2008deliberative}
Sarah Michaels, Catherine O'Connor, and Lauren~B. Resnick.
\newblock Deliberative discourse idealized and realized: Accountable talk in the classroom and in civic life.
\newblock \emph{Studies in Philosophy and Education}, 27\penalty0 (4):\penalty0 283--297, 2008.
\newblock \doi{10.1007/s11217-007-9071-1}.

\bibitem[Na \& Feng(2025)Na and Feng]{na2025llm}
Ying Na and Shihui Feng.
\newblock Llm-assisted automated deductive coding of dialogue data: leveraging dialogue-specific characteristics to enhance contextual understanding.
\newblock In \emph{International Conference on Artificial Intelligence in Education}, pp.\  248--262. Springer, 2025.

\bibitem[Naim et~al.(2025)Naim, Cao, Tasneem, Jacobs, Milne, Martin, and Sumner]{naim2025actionable}
Jannatun Naim, Jie Cao, Fareen Tasneem, Jennifer Jacobs, Brent Milne, James Martin, and Tamara Sumner.
\newblock Towards actionable pedagogical feedback: A multi-perspective analysis of mathematics teaching and tutoring dialogue.
\newblock In \emph{Proceedings of the 18th International Conference on Educational Data Mining}, 2025.

\bibitem[Ostyakova et~al.(2023)Ostyakova, Smilga, Petukhova, Molchanova, and Kornev]{ostyakova2023chatgpt}
Lidiia Ostyakova, Veronika Smilga, Kseniia Petukhova, Maria Molchanova, and Daniel Kornev.
\newblock Chatgpt vs. crowdsourcing vs. experts: Annotating open-domain conversations with speech functions.
\newblock In \emph{Proceedings of the 24th Annual Meeting of the Special Interest Group on Discourse and Dialogue}, pp.\  242--254, 2023.

\bibitem[Qamar et~al.(2025)Qamar, Tong, and Huang]{qamar-etal-2025-llms}
Ayesha Qamar, Jonathan Tong, and Ruihong Huang.
\newblock Do {LLM}s understand dialogues? a case study on dialogue acts.
\newblock In \emph{Proceedings of the 63rd Annual Meeting of the Association for Computational Linguistics (Volume 1: Long Papers)}, pp.\  26219--26237, Vienna, Austria, jul 2025. Association for Computational Linguistics.
\newblock \doi{10.18653/v1/2025.acl-long.1271}.
\newblock URL \url{https://aclanthology.org/2025.acl-long.1271/}.

\bibitem[Qi et~al.(2024)Qi, Zheng, Wei, Xu, Chen, and Gu]{qi2024edudcm}
Changyong Qi, Longwei Zheng, Yuang Wei, Haoxin Xu, Peiji Chen, and Xiaoqing Gu.
\newblock Edudcm: a novel framework for automatic educational dialogue classification dataset construction via distant supervision and large language models.
\newblock \emph{Applied Sciences}, 15\penalty0 (1):\penalty0 154, 2024.

\bibitem[Scarlatos et~al.(2025)Scarlatos, Baker, and Lan]{scarlatos2025exploring}
Alexander Scarlatos, Ryan~S. Baker, and Andrew Lan.
\newblock Exploring knowledge tracing in tutor-student dialogues using llms.
\newblock In \emph{Proceedings of the 15th International Learning Analytics and Knowledge Conference}, pp.\  1--10, 2025.

\bibitem[Shah et~al.(2025)Shah, White, Andersen, Buscher, Counts, Das, Montazer, Manivannan, Neville, Rangan, et~al.]{shah2025using}
Chirag Shah, Ryen White, Reid Andersen, Georg Buscher, Scott Counts, Sarkar Das, Ali Montazer, Sathish Manivannan, Jennifer Neville, Nagu Rangan, et~al.
\newblock Using large language models to generate, validate, and apply user intent taxonomies.
\newblock \emph{ACM Transactions on the Web}, 19\penalty0 (3):\penalty0 1--29, 2025.

\bibitem[Shan et~al.(2023)Shan, Wang, Zhang, Kao, and Chan]{shan2023annotating}
Dou Shan, Dong Wang, Chen Zhang, Kimberly~Britt Kao, and Carol Ka~Yuk Chan.
\newblock Annotating educational dialog act with data augmentation in online one-on-one tutoring.
\newblock In \emph{Proceedings of the 24th International Conference on Artificial Intelligence in Education}, pp.\  473--486, 2023.

\bibitem[Su \& Ye(2025)Su and Ye]{su2025large}
Hang Su and Jun Ye.
\newblock Large language models for automating fine-grained speech act annotation: A critical evaluation of gpt-4o and deepseek.
\newblock \emph{Corpus Pragmatics}, pp.\  1--20, 2025.

\bibitem[Suresh et~al.(2022)Suresh, Jacobs, Harty, Perkoff, Martin, and Sumner]{suresh2022talkmoves}
Abhijit Suresh, Jennifer Jacobs, Charis Harty, Margaret Perkoff, James~H. Martin, and Tamara Sumner.
\newblock The talkmoves dataset: K-12 mathematics lesson transcripts annotated for teacher and student discursive moves.
\newblock In \emph{Proceedings of the Thirteenth Language Resources and Evaluation Conference}, pp.\  4654--4662, Marseille, France, June 2022. European Language Resources Association.
\newblock URL \url{https://aclanthology.org/2022.lrec-1.497/}.

\bibitem[Tan et~al.(2023)]{tan2023informativeness}
Wei Tan et~al.
\newblock Does informativeness matter? active learning for educational dialogue act classification.
\newblock In \emph{Proceedings of the 24th International Conference on Artificial Intelligence in Education}, pp.\  115--127, 2023.

\bibitem[Tavakoli \& Zamani(2025)Tavakoli and Zamani]{10.1145/3731120.3744574}
Leila Tavakoli and Hamed Zamani.
\newblock Reliable annotations with less effort: Evaluating llm-human collaboration in search clarifications.
\newblock In \emph{Proceedings of the 2025 International ACM SIGIR Conference on Innovative Concepts and Theories in Information Retrieval (ICTIR)}, ICTIR '25, pp.\  92--102. Association for Computing Machinery, 2025.
\newblock \doi{10.1145/3731120.3744574}.

\bibitem[Vail \& Boyer(2014)Vail and Boyer]{vail2014refinement}
Andrew~K. Vail and Kristy~Elizabeth Boyer.
\newblock Identifying effective moves in tutoring: On the refinement of dialogue act annotation schemes.
\newblock In \emph{Proceedings of the 12th International Conference on Intelligent Tutoring Systems}, pp.\  199--209, 2014.

\bibitem[Wang et~al.(2025)Wang, Gao, Xiao, et~al.]{wang2025segmentation}
Zhitong Wang, Cheng Gao, Chaojun Xiao, et~al.
\newblock Document segmentation matters for retrieval-augmented generation.
\newblock In \emph{Findings of the Association for Computational Linguistics: ACL 2025}, 2025.

\bibitem[Xiao et~al.(2023)Xiao, Liu, Zhang, and Muennighoff]{xiao2023cpack}
Shitao Xiao, Zheng Liu, Peitian Zhang, and Niklas Muennighoff.
\newblock C-pack: Packaged resources to advance general chinese embedding.
\newblock \emph{arXiv preprint arXiv:2309.07597}, 2023.

\bibitem[Xu et~al.(2025)Xu, Liu, Nag, Dai, Xie, Tang, Luo, Li, Ho, Yang, and He]{xu2025simrag}
Ran Xu, Hui Liu, Sreyashi Nag, Zhenwei Dai, Yaochen Xie, Xianfeng Tang, Chen Luo, Yang Li, Joyce~C. Ho, Carl Yang, and Qi~He.
\newblock Simrag: Self-improving retrieval-augmented generation for adapting large language models to specialized domains.
\newblock In \emph{Proceedings of the 2025 Conference of the North American Chapter of the Association for Computational Linguistics}, 2025.

\bibitem[Yang et~al.(2023)Yang, Christmann, and Gasevic]{yang2023robust}
Linjuan Yang, Philipp Christmann, and Dragan Gasevic.
\newblock Robust educational dialogue act classifiers with low-resource and imbalanced datasets.
\newblock In \emph{Proceedings of the 24th International Conference on Artificial Intelligence in Education}, pp.\  114--126, 2023.

\bibitem[Yu et~al.(2024)Yu, Li, Su, and Fuoli]{yu2024assessing}
Danni Yu, Luyang Li, Hang Su, and Matteo Fuoli.
\newblock Assessing the potential of llm-assisted annotation for corpus-based pragmatics and discourse analysis: The case of apology.
\newblock \emph{International Journal of Corpus Linguistics}, 29\penalty0 (4):\penalty0 534--561, 2024.

\bibitem[Zent et~al.(2025)Zent, Smith, and Woodhead]{zent2025eedi}
Matthew Zent, Digory Smith, and Simon Woodhead.
\newblock Question-anchored tutoring dialogues.
\newblock \url{https://huggingface.co/datasets/Eedi/Question-Anchored-Tutoring-Dialogues-2k}, 2025.
\newblock Eedi dataset. Accessed: 2026-03-29.

\end{thebibliography}
\bibliographystyle{colm2026_conference}
\clearpage
\appendix
\section{Prompt Templates}
\label{app:prompts}
This section presents the full annotation prompt used in our pipeline. The prompt consists of a system message defining the annotator role, codebook, and output format, followed by a user message presenting retrieved examples and the target dialogue context.
\subsection{System Message}
\begin{quote}
\small
You are an expert educational discourse analyst. Your task is to label each teacher utterance with one move from the allowed moves list. \\
\textbf{Workflow}
\begin{enumerate}
    \item Read the dialogue carefully.
    \item For each teacher utterance, assign exactly ONE Move from the Allowed Moves list.
    \item If an utterance could fit multiple moves, choose the one that best represents the communicative function in context.
    \item If no Move applies (e.g., the utterance is off-topic, evaluative, unclear, or unrelated to the lesson), do not label it and leave it as None.
\end{enumerate}
\textbf{Allowed Moves \& Definitions}
\begin{itemize}
    \item \textbf{Keeping Everyone Together} -- Teacher prompts students to be active listeners and orienting students to each other.
    \item \textbf{Getting Students to Relate to Another's Ideas} -- Teacher prompts students to react to what a classmate said.
    \item \textbf{Restating} -- Teacher repeats all or part of what a student said word for word.
    \item \textbf{Pressing for Accuracy} -- Teacher prompts students to make a mathematical contribution or use mathematical language.
    \item \textbf{Revoicing} -- Teacher repeats what a student said but adding on or changing the wording.
    \item \textbf{Pressing for Reasoning} -- Teacher prompts students to explain, provide evidence, share their thinking behind a decision, or connect ideas or representations.
\end{itemize}
\textbf{Rules}
\begin{itemize}
    \item Annotate ONLY the utterance marked with \texttt{<<<TARGET>>>}.
    \item Consider the surrounding context to understand the utterance's function.
    \item If example annotations from similar dialogues are provided, use them as reference.
    \item Respond with ONLY a JSON object: \texttt{\{"label": "<label or null>", "confidence": <0.0-1.0>\}}
    \item If no Move applies, use \texttt{null} for label.
    \item Do not include any other text outside the JSON.
\end{itemize}
\end{quote}
\subsection{User Message}
For retrieval-augmented conditions, the user message begins with $k$ retrieved examples, each showing the chunk text and its majority label. For \textsc{no\_rag}, this section is omitted.
\begin{quote}
\small
\textbf{Similar annotated examples from the corpus}
\\
\texttt{Example 1 (labeled: <majority\_label>)}\\
\texttt{<chunk text>}
\texttt{[... up to k examples ...]} \\
\textbf{Current dialogue context}
\texttt{<dynamic\_context>}
Annotate the utterance marked \texttt{<<<TARGET>>>} by \texttt{[<target\_speaker>]}.\\
Respond with only the JSON object.
\end{quote}
The \texttt{<dynamic\_context>} field contains the context window from Section~\ref{sec:query}, with the target utterance marked by \texttt{<<<TARGET>>>}. Each utterance is formatted as \texttt{[speaker] utterance text}.
\section{Additional Results}
\label{app:results}
\subsection{Per-Label Cohen's $\kappa$ (GPT-5.2)}
Tables~\ref{tab:perlabel_tm} and~\ref{tab:perlabel_eedi} report per-label Cohen's $\kappa$ for all three conditions with GPT-5.2, with labels ordered by frequency.
\begin{table}[h]
\centering
\small
\begin{tabular}{lcccccc}
\toprule
 & ACC & KET & REV & RES & REA & GSR \\
\midrule
\textsc{no\_rag} & 0.250 & 0.472 & 0.205 & 0.198 & 0.232 & 0.367 \\
\textsc{RAG\_no\_finetune} & 0.473 & 0.562 & 0.362 & 0.386 & 0.416 & 0.459 \\
\textsc{RAG\_finetuned\_utt} & \textbf{0.552} & \textbf{0.591} & \textbf{0.442} & \textbf{0.479} & \textbf{0.445} & \textbf{0.540} \\
\midrule
Support ($n$) & 1,921 & 1,829 & 367 & 288 & 198 & 145 \\
\bottomrule
\end{tabular}
\caption{Per-label Cohen's $\kappa$ on TalkMoves test ($k = 3$, GPT-5.2).}
\label{tab:perlabel_tm}
\end{table}
\begin{table}[h]
\centering
\small
\begin{tabular}{lcccccc}
\toprule
 & ACC & KET & REV & REA & GSR & RES \\
\midrule
\textsc{no\_rag} & 0.344 & 0.584 & 0.334 & 0.104 & 0.106 & 0.006 \\
\textsc{RAG\_no\_finetune} & 0.551 & 0.671 & 0.263 & 0.250 & 0.213 & 0.020 \\
\textsc{RAG\_finetuned\_utt} & \textbf{0.662} & \textbf{0.739} & \textbf{0.642} & \textbf{0.380} & \textbf{0.418} & \textbf{0.444} \\
\midrule
Support ($n$) & 5,286 & 2,779 & 1,531 & 157 & 81 & 9 \\
\bottomrule
\end{tabular}
\caption{Per-label Cohen's $\kappa$ on Eedi test ($k = 3$, GPT-5.2). RES has only 9 support instances. \textsc{RAG\_no\_finetune} underperforms \textsc{no\_rag} on REV ($0.263$ vs.\ $0.334$).}
\label{tab:perlabel_eedi}
\end{table}
\subsection{Classification Reports (GPT-5.2)}
\begin{table}[htbp!]
\centering
\small
\begin{tabular}{lcccc}
\toprule
 & Precision & Recall & F1 & Support \\
\midrule
KET & 0.68 & 0.66 & 0.67 & 1,829 \\
GSR & 0.55 & 0.29 & 0.38 & 145 \\
RES & 0.22 & 0.30 & 0.25 & 287 \\
ACC & 0.62 & 0.41 & 0.49 & 1,921 \\
REV & 0.56 & 0.14 & 0.23 & 366 \\
REA & 0.17 & 0.96 & 0.29 & 198 \\
\midrule
Accuracy & & & 0.50 & 4,746 \\
Macro avg & 0.47 & 0.46 & 0.39 & 4,746 \\
Weighted avg & 0.59 & 0.50 & 0.52 & 4,746 \\
\midrule
\multicolumn{4}{l}{Cohen's $\kappa = 0.315$} \\
\bottomrule
\end{tabular}
\caption{\textsc{no\_rag} classification report on TalkMoves test (GPT-5.2).}
\label{tab:clf_norag_tm}
\end{table}
\begin{table}[h]
\centering
\small
\begin{tabular}{lcccc}
\toprule
 & Precision & Recall & F1 & Support \\
\midrule
KET & 0.77 & 0.66 & 0.71 & 701 \\
GSR & 0.70 & 0.36 & 0.47 & 73 \\
RES & 0.46 & 0.39 & 0.42 & 111 \\
ACC & 0.68 & 0.71 & 0.69 & 774 \\
REV & 0.61 & 0.29 & 0.39 & 147 \\
REA & 0.30 & 0.99 & 0.46 & 84 \\
\midrule
Accuracy & & & 0.64 & 1,890 \\
Macro avg & 0.59 & 0.57 & 0.52 & 1,890 \\
Weighted avg & 0.68 & 0.64 & 0.64 & 1,890 \\
\midrule
\multicolumn{4}{l}{Cohen's $\kappa = 0.479$} \\
\bottomrule
\end{tabular}
\caption{\textsc{RAG\_no\_finetune} classification report on TalkMoves test (GPT-5.2).}
\label{tab:clf_raggeneral_tm}
\end{table}
\begin{table}[htbp!]
\centering
\small
\begin{tabular}{lcccc}
\toprule
 & Precision & Recall & F1 & Support \\
\midrule
KET & 0.83 & 0.65 & 0.73 & 534 \\
GSR & 0.64 & 0.49 & 0.55 & 43 \\
RES & 0.53 & 0.48 & 0.50 & 69 \\
ACC & 0.72 & 0.78 & 0.75 & 589 \\
REV & 0.68 & 0.36 & 0.47 & 90 \\
REA & 0.32 & 0.97 & 0.48 & 62 \\
\midrule
Accuracy & & & 0.69 & 1,387 \\
Macro avg & 0.62 & 0.62 & 0.58 & 1,387 \\
Weighted avg & 0.73 & 0.69 & 0.69 & 1,387 \\
\midrule
\multicolumn{4}{l}{Cohen's $\kappa = 0.539$} \\
\bottomrule
\end{tabular}
\caption{\textsc{RAG\_finetuned\_utt} classification report on TalkMoves test (GPT-5.2).}
\label{tab:clf_ragfinetuned_tm}
\end{table}
\begin{table}[h]
\centering
\small
\begin{tabular}{lcccc}
\toprule
 & Precision & Recall & F1 & Support \\
\midrule
KET & 0.79 & 0.59 & 0.67 & 2,335 \\
GSR & 0.21 & 0.07 & 0.11 & 81 \\
RES & 0.00 & 0.17 & 0.01 & 6 \\
ACC & 0.75 & 0.65 & 0.69 & 5,265 \\
REV & 0.95 & 0.23 & 0.37 & 1,334 \\
REA & 0.07 & 0.99 & 0.13 & 157 \\
\midrule
Accuracy & & & 0.57 & 9,178 \\
Macro avg & 0.46 & 0.45 & 0.33 & 9,178 \\
Weighted avg & 0.77 & 0.57 & 0.63 & 9,178 \\
\midrule
\multicolumn{4}{l}{Cohen's $\kappa = 0.351$} \\
\bottomrule
\end{tabular}
\caption{\textsc{no\_rag} classification report on Eedi test (GPT-5.2).}
\label{tab:clf_norag_eedi}
\end{table}
\begin{table}[h]
\centering
\small
\begin{tabular}{lcccc}
\toprule
 & Precision & Recall & F1 & Support \\
\midrule
KET & 0.86 & 0.67 & 0.75 & 1,788 \\
GSR & 0.40 & 0.15 & 0.22 & 54 \\
RES & 0.01 & 0.50 & 0.02 & 4 \\
ACC & 0.78 & 0.86 & 0.81 & 3,522 \\
REV & 0.94 & 0.17 & 0.29 & 842 \\
REA & 0.16 & 1.00 & 0.27 & 105 \\
\midrule
Accuracy & & & 0.71 & 6,315 \\
Macro avg & 0.52 & 0.56 & 0.39 & 6,315 \\
Weighted avg & 0.81 & 0.71 & 0.71 & 6,315 \\
\midrule
\multicolumn{4}{l}{Cohen's $\kappa = 0.505$} \\
\bottomrule
\end{tabular}
\caption{\textsc{RAG\_no\_finetune} classification report on Eedi test (GPT-5.2).}
\label{tab:clf_raggeneral_eedi}
\end{table}
\begin{table}[h]
\centering
\small
\begin{tabular}{lcccc}
\toprule
 & Precision & Recall & F1 & Support \\
\midrule
KET & 0.96 & 0.69 & 0.80 & 852 \\
GSR & 0.44 & 0.41 & 0.42 & 27 \\
RES & 0.29 & 1.00 & 0.44 & 2 \\
ACC & 0.82 & 0.93 & 0.87 & 1,753 \\
REV & 0.95 & 0.52 & 0.67 & 359 \\
REA & 0.25 & 1.00 & 0.40 & 52 \\
\midrule
Accuracy & & & 0.81 & 3,045 \\
Macro avg & 0.62 & 0.76 & 0.60 & 3,045 \\
Weighted avg & 0.86 & 0.81 & 0.81 & 3,045 \\
\midrule
\multicolumn{4}{l}{Cohen's $\kappa = 0.659$} \\
\bottomrule
\end{tabular}
\caption{\textsc{RAG\_finetuned\_utt} classification report on Eedi test (GPT-5.2).}
\label{tab:clf_ragfinetuned_eedi}
\end{table}
\section{Model Comparison Per-Label Results}
\label{app:model_comparison}
Tables~\ref{tab:model_perlabel_tm} and~\ref{tab:model_perlabel_eedi} report per-label Cohen's $\kappa$ for all three LLM backbones under each retrieval condition ($k = 3$).
\begin{table}[htbp!]
\centering
\small
\begin{tabular}{llcccccc}
\toprule
Model & Condition & KET & GSR & RES & ACC & REV & REA \\
\midrule
GPT-5.2 & \textsc{no\_rag} & 0.472 & 0.367 & 0.198 & 0.250 & 0.205 & 0.232 \\
GPT-5.2 & \textsc{RAG\_no\_finetune} & 0.562 & 0.459 & 0.386 & 0.473 & 0.362 & 0.416 \\
GPT-5.2 & \textsc{RAG\_finetuned\_utt} & 0.591 & 0.540 & 0.479 & 0.552 & 0.442 & 0.445 \\
\midrule
Sonnet 4.6 & \textsc{no\_rag} & 0.421 & 0.506 & 0.410 & 0.397 & 0.548 & 0.353 \\
Sonnet 4.6 & \textsc{RAG\_no\_finetune} & 0.577 & 0.464 & 0.461 & 0.480 & 0.522 & 0.360 \\
Sonnet 4.6 & \textsc{RAG\_finetuned\_utt} & \textbf{0.631} & 0.528 & \textbf{0.542} & \textbf{0.594} & \textbf{0.566} & \textbf{0.448} \\
\midrule
Qwen3-32b & \textsc{no\_rag} & 0.393 & 0.406 & 0.366 & 0.090 & 0.630 & 0.224 \\
Qwen3-32b & \textsc{RAG\_no\_finetune} & 0.490 & 0.327 & 0.380 & 0.440 & 0.359 & 0.427 \\
Qwen3-32b & \textsc{RAG\_finetuned\_utt} & 0.584 & 0.411 & 0.367 & 0.584 & 0.405 & 0.426 \\
\bottomrule
\end{tabular}
\caption{Per-label Cohen's $\kappa$ on TalkMoves test across three LLM backbones ($k = 3$). Bold indicates the best \textsc{RAG\_finetuned\_utt} result per label.}
\label{tab:model_perlabel_tm}
\end{table}
\begin{table}[htbp!]
\centering
\small
\begin{tabular}{llcccccc}
\toprule
Model & Condition & KET & GSR & RES & ACC & REV & REA \\
\midrule
GPT-5.2 & \textsc{no\_rag} & 0.584 & 0.106 & 0.006 & 0.344 & 0.334 & 0.104 \\
GPT-5.2 & \textsc{RAG\_no\_finetune} & 0.671 & 0.213 & 0.020 & 0.551 & 0.263 & 0.250 \\
GPT-5.2 & \textsc{RAG\_finetuned\_utt} & 0.739 & \textbf{0.418} & 0.444$^\dagger$ & 0.662 & 0.642 & 0.380 \\
\midrule
Sonnet 4.6 & \textsc{no\_rag} & 0.303 & 0.300 & --- & 0.478 & 0.643 & 0.207 \\
Sonnet 4.6 & \textsc{RAG\_no\_finetune} & 0.683 & 0.292 & --- & 0.670 & 0.653 & 0.330 \\
Sonnet 4.6 & \textsc{RAG\_finetuned\_utt} & 0.749 & 0.387 & 0.104 & \textbf{0.775} & \textbf{0.855} & 0.361 \\
\midrule
Qwen3-32b & \textsc{no\_rag} & 0.201 & 0.275 & --- & 0.129 & 0.571 & 0.088 \\
Qwen3-32b & \textsc{RAG\_no\_finetune} & 0.690 & 0.273 & --- & 0.630 & 0.356 & 0.443 \\
Qwen3-32b & \textsc{RAG\_finetuned\_utt} & \textbf{0.764} & 0.368 & 0.092 & 0.734 & 0.665 & \textbf{0.400} \\
\bottomrule
\end{tabular}
\caption{Per-label Cohen's $\kappa$ on Eedi test across three LLM backbones ($k = 3$). Bold indicates the best \textsc{RAG\_finetuned\_utt} result per label. $^\dagger$RES has only 9 support instances}
\label{tab:model_perlabel_eedi}
\end{table}
\section{Fixed-Window Chunking Ablation}
\label{app:fixedwindow}
To isolate the contribution of semantic boundary detection from other components of the pipeline, we compare our calibrated semantic chunking against naive fixed-window baselines. Fixed-window chunking partitions each dialogue into non-overlapping segments of exactly $w$ utterances, with no boundary detection or similarity computation. We test $w = 3$ and $w = 5$, both using Cohere embeddings for chunk-level retrieval to match the \textsc{RAG\_no\_finetune} condition.
\subsection{Overall Results}
\begin{table}[h]
\centering
\small
\begin{tabular}{llcccc}
\toprule
Condition & Chunking & Retrieval & TM $\kappa$ & Eedi $\kappa$ \\
\midrule
\textsc{no\_rag} & None & None & 0.313 & 0.351 \\
\textsc{RAG\_no\_finetune} & Semantic (Cohere) & Cohere & 0.474 & 0.505 \\
\textsc{rag\_fixed\_w3} & Fixed ($w{=}3$) & Cohere & 0.500 & 0.571 \\
\textsc{rag\_fixed\_w5} & Fixed ($w{=}5$) & Cohere & 0.475 & 0.519 \\
\midrule
\textsc{RAG\_finetuned\_utt} & Fine-tuned BGE & Fine-tuned BGE & \textbf{0.539} & \textbf{0.659} \\
\bottomrule
\end{tabular}
\caption{Fixed-window vs. semantic chunking ablation on test sets ($k = 3$, GPT-5.2). All conditions except \textsc{no\_rag} and \textsc{RAG\_finetuned\_utt} use Cohere for retrieval. Sample sizes differ across conditions due to index coverage.}
\label{tab:fixedwindow}
\end{table}
Table~\ref{tab:fixedwindow} presents Cohen's $\kappa$ across all chunking conditions on both test sets with $k = 3$ and GPT-5.2. Several observations emerge. First, fixed-window chunking with $w = 3$ matches or slightly outperforms semantic chunking when the same retrieval model (Cohere) is used. On TalkMoves, \textsc{rag\_fixed\_w3} ($\kappa = 0.500$) exceeds \textsc{RAG\_no\_finetune} ($0.474$). On Eedi, the same pattern holds ($0.571$ vs. $0.505$). This suggests that for chunk-level retrieval with general-purpose embeddings, shorter fixed windows may actually produce better retrieval units than semantic chunking, possibly because smaller chunks reduce the dilution effect discussed in Section~\ref{sec:index}.
Second, $w = 3$ consistently outperforms $w = 5$ on both datasets, reinforcing that shorter chunks produce more focused retrieval matches when using chunk-level indexing.
Third, and most importantly, \textsc{RAG\_finetuned\_utt} outperforms all other conditions by a substantial margin despite using semantic rather than fixed-window chunking. This confirms that the dominant performance driver is the combination of domain-adapted embeddings with utterance-level indexing, not the chunking strategy itself. The chunking strategy matters at the margin when all else is held equal, but the choice of embedding model and index granularity accounts for the majority of the performance gap.
\subsection{Per-Label $\kappa$ for Fixed-Window Conditions}
\begin{table}[htbp!]
\centering
\small
\begin{tabular}{lcccccc}
\toprule
 & KET & GSR & RES & ACC & REV & REA \\
\midrule
\textsc{rag\_fixed\_w3} & 0.554 & 0.254 & \textbf{0.731} & 0.470 & \textbf{0.549} & 0.350 \\
\textsc{rag\_fixed\_w5} & 0.537 & 0.396 & 0.691 & 0.447 & 0.553 & 0.293 \\
\midrule
\textsc{RAG\_finetuned\_utt} & \textbf{0.591} & \textbf{0.540} & 0.479 & \textbf{0.552} & 0.442 & \textbf{0.445} \\
\bottomrule
\end{tabular}
\caption{Per-label $\kappa$ on TalkMoves test for fixed-window conditions vs. \textsc{RAG\_finetuned\_utt}.}
\label{tab:fixedwindow_perlabel_tm}
\end{table}
\begin{table}[htbp!]
\centering
\small
\begin{tabular}{lcccccc}
\toprule
 & KET & GSR & RES & ACC & REV & REA \\
\midrule
\textsc{rag\_fixed\_w3} & 0.676 & 0.195 & -0.000 & 0.599 & 0.534 & 0.256 \\
\textsc{rag\_fixed\_w5} & 0.607 & 0.069 & 0.166 & 0.552 & 0.592 & 0.187 \\
\midrule
\textsc{RAG\_finetuned\_utt} & \textbf{0.739} & \textbf{0.418} & \textbf{0.444} & \textbf{0.662} & \textbf{0.642} & \textbf{0.380} \\
\bottomrule
\end{tabular}
\caption{Per-label $\kappa$ on Eedi test for fixed-window conditions vs. \textsc{RAG\_finetuned\_utt}.}
\label{tab:fixedwindow_perlabel_eedi}
\end{table}
Tables~\ref{tab:fixedwindow_perlabel_tm} and~\ref{tab:fixedwindow_perlabel_eedi} break down the fixed-window results by label. An interesting pattern is that \textsc{rag\_fixed\_w3} achieves notably high $\kappa$ on RES (0.731) and REV (0.549) on TalkMoves, outperforming even \textsc{RAG\_finetuned\_utt} on those two labels. This may reflect that Restating and Revoicing are highly localized acts (a tutor repeating or rephrasing what a student just said), making short fixed windows effective retrieval units for these specific categories. However, \textsc{RAG\_finetuned\_utt} achieves more balanced performance across all labels and substantially higher $\kappa$ on the labels that require broader contextual understanding (GSR, ACC, REA). On Eedi, \textsc{RAG\_finetuned\_utt} dominates on every label.
\section{Retrieval Label Match Rate}
\label{app:retrieval_quality}
To directly evaluate whether domain-adapted embeddings improve retrieval quality independently of LLM annotation, we measure the label match rate between retrieved demonstrations and the target utterance's gold label. For each target utterance, we check whether the top-1 retrieved example shares the same gold label (Top-1) and whether any of the top-$k$ retrieved examples share the label (Any-$k$, with $k = 3$).

Domain-adapted retrieval produces a 14 to 22 percentage point improvement in label match rate across both datasets and both metrics. The gains are especially pronounced for labels that are rare or context-dependent. On TalkMoves, the any-$k$ match rate for ACC jumps from 29\% to 79\% with fine-tuned embeddings, directly explaining the large annotation $\kappa$ improvement for that label. KET, the most frequent label, achieves near-perfect retrieval (96--97\% any-$k$) under both conditions, consistent with its high annotation $\kappa$ across all models.
Conversely, labels where retrieval quality remains low correspond to persistently low annotation performance. RES on Eedi has a 0\% match rate for both conditions, reflecting its extreme rarity in the training index (9 labeled instances). Neither embedding model can retrieve relevant RES demonstrations, which directly explains the near-zero $\kappa$ for this label. GSR and REV both remain below 45\% any-$k$ match rate, consistent with their lower support and higher chunk impurity. These patterns confirm that retrieval precision is the mechanistic bottleneck for annotation quality and that the pipeline degrades predictably for labels that are underrepresented in the training index.

\section{Confidence Calibration Analysis}
\label{app:confidence}
Our pipeline produces a confidence score alongside each predicted label. We analyze whether these scores are informative by examining (1) how confidence distributions shift across retrieval conditions and (2) whether high-confidence predictions are reliably correct.
\begin{figure}[htpb!]
    \centering
    \includegraphics[width=\textwidth]{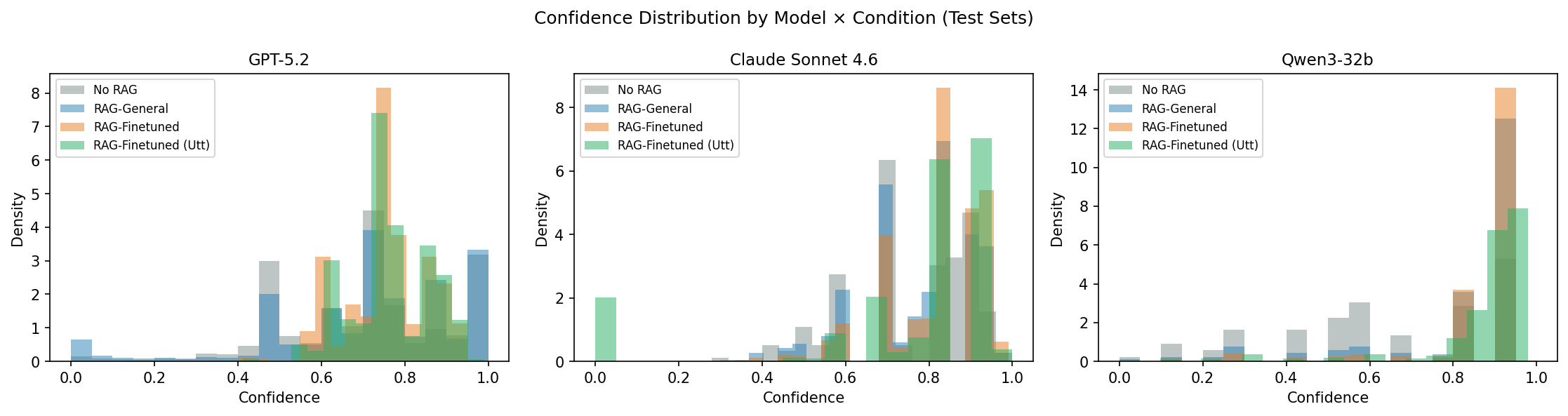}
    \caption{Confidence score distributions by model and retrieval condition on the combined test sets. Domain-adapted retrieval (\textsc{RAG\_finetuned\_utt}) concentrates prediction mass at higher confidence levels across all three models.}
    \label{fig:confidence}
\end{figure}
Figure~\ref{fig:confidence} shows that retrieval shifts the confidence distribution rightward for all three models. With \textsc{RAG\_finetuned\_utt}, predictions concentrate in the 0.8--1.0 range, consistent with the high retrieval label match rates reported in Appendix~\ref{app:retrieval_quality}. Without retrieval, confidence scores are more dispersed, and for GPT-5.2 a notable mass appears near zero, indicating frequent uncertainty.
To assess whether confidence scores support reliable triage, we compute Cohen's $\kappa$ and coverage on the subset of predictions with confidence $\geq 0.9$ (Table~\ref{tab:confidence}).
\begin{table}[h]
\centering
\small
\begin{tabular}{llcc}
\toprule
Model & Condition & $\kappa$ ($\geq 0.9$) & Coverage \\
\midrule
GPT-5.2 & \textsc{no\_rag} & 0.086 & 19.1\% \\
GPT-5.2 & \textsc{RAG\_no\_finetune} & 0.412 & 23.8\% \\
GPT-5.2 & \textsc{RAG\_finetuned\_utt} & \textbf{0.939} & 11.6\% \\
\midrule
Sonnet 4.6 & \textsc{no\_rag} & 0.594 & 3.9\% \\
Sonnet 4.6 & \textsc{RAG\_no\_finetune} & 0.838 & 24.3\% \\
Sonnet 4.6 & \textsc{RAG\_finetuned\_utt} & 0.896 & 34.3\% \\
\midrule
Qwen3-32b & \textsc{no\_rag} & 0.212 & 17.0\% \\
Qwen3-32b & \textsc{RAG\_no\_finetune} & 0.611 & 59.8\% \\
Qwen3-32b & \textsc{RAG\_finetuned\_utt} & 0.685 & 68.3\% \\
\bottomrule
\end{tabular}
\caption{Cohen's $\kappa$ and coverage for predictions with confidence $\geq 0.9$. Coverage is the fraction of test utterances where the model's confidence exceeds the threshold. GPT-5.2 with \textsc{RAG\_finetuned\_utt} achieves near-perfect agreement ($\kappa = 0.939$) on its high-confidence subset, though coverage is limited to 11.6\%.}
\label{tab:confidence}
\end{table}
Two patterns emerge. First, \textsc{RAG\_finetuned\_utt} substantially improves the reliability of high-confidence predictions across all models. GPT-5.2 with \textsc{RAG\_finetuned\_utt} achieves $\kappa = 0.939$ on its confident subset, compared to $0.086$ without retrieval, indicating that domain-adapted retrieval nearly eliminates errors among predictions the model is most certain about. Second, models differ in how they distribute confidence. Sonnet 4.6 with \textsc{RAG\_finetuned\_utt} achieves both high $\kappa$ ($0.896$) and high coverage ($34.3\%$), making it the strongest candidate for a human-in-the-loop workflow where confident predictions are auto-accepted and uncertain ones are flagged for manual review. Qwen3-32b produces the highest coverage ($68.3\%$) but at lower $\kappa$ ($0.685$), suggesting it is overconfident relative to its actual accuracy. These results support using confidence scores as a practical triage mechanism in deployment, though the coverage-accuracy tradeoff varies across models and should be calibrated for specific use cases.
\section{Effect of Retrieval Depth}
\label{app:k_ablation}
\begin{figure}[htpb!]
    \centering
    \includegraphics[width=0.8\textwidth]{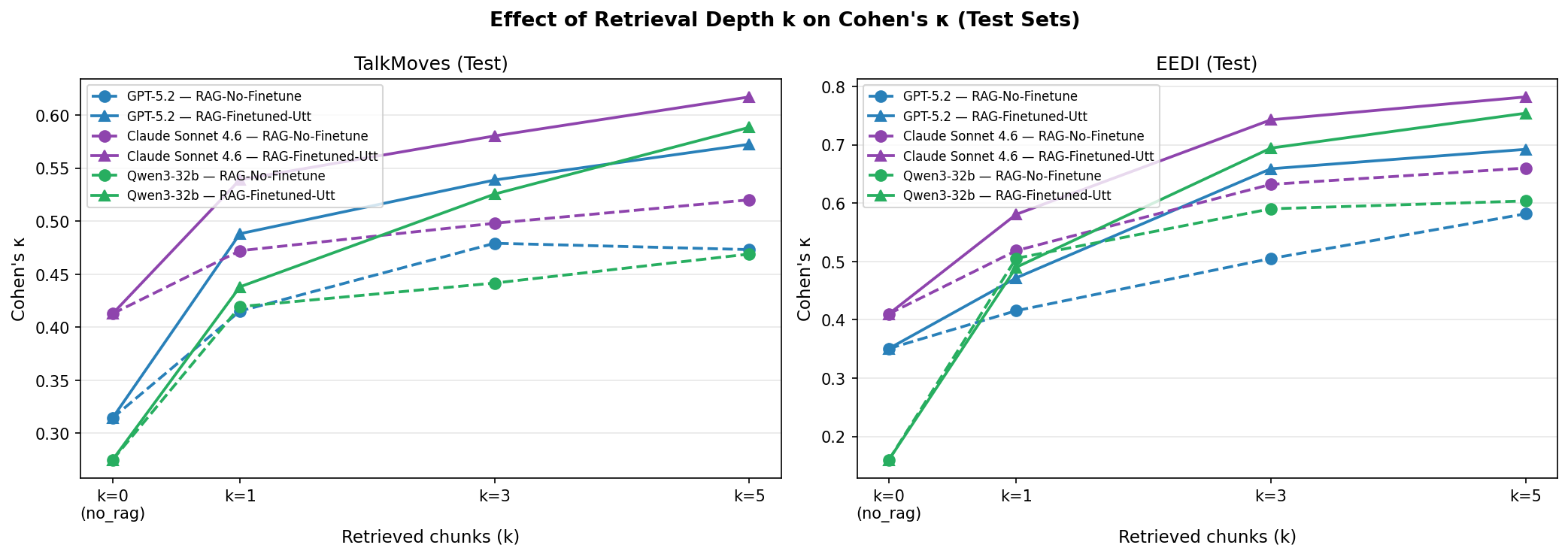}
    \caption{Cohen's $\kappa$ as a function of retrieval depth $k$ across three LLM backbones on TalkMoves test (top) and Eedi test (bottom). Left panels show \textsc{RAG\_no\_finetune}, right panels show \textsc{RAG\_finetuned\_utt}. Domain-adapted retrieval consistently outperforms general-purpose retrieval at every $k$ for all models.}
    \label{fig:k_ablation}
\end{figure}
Figure~\ref{fig:k_ablation} shows annotation quality as a function of retrieval depth $k$ across all three models. \textsc{RAG\_finetuned\_utt} outperforms \textsc{RAG\_no\_finetune} at every value of $k$ for all models, confirming that domain-adapted retrieval consistently produces more informative demonstrations regardless of the LLM backbone. A single retrieved example ($k = 1$) is clearly insufficient across all conditions. Performance continues to improve through $k = 5$ on Eedi for all models, with Sonnet 4.6 reaching $\kappa = 0.782$. On TalkMoves, $\kappa$ also improves through $k = 5$ under \textsc{RAG\_finetuned\_utt} for all models, while \textsc{RAG\_no\_finetune} plateaus or slightly declines after $k = 3$, possibly because additional chunks from a general-purpose index introduce noise. The widening gap between \textsc{RAG\_finetuned\_utt} and \textsc{RAG\_no\_finetune} as $k$ increases suggests that domain-adapted embeddings become increasingly important as retrieval depth grows.

\section{Confound Isolation Analysis}
\label{app:confound}
The comparison between \textsc{RAG\_no\_finetune} and \textsc{RAG\_finetuned\_utt} conflates two changes: embedding model (Cohere vs. fine-tuned BGE) and index granularity (chunk vs. utterance). To disentangle these factors, we evaluate \textsc{RAG\_finetuned\_chunk}, which uses the fine-tuned BGE model with chunk-level indexing. Each step changes exactly one variable, and we report results for all three LLM backbones (Tables~\ref{tab:confound_tm} and~\ref{tab:confound_eedi}).
\begin{table}[h]
\centering
\small
\begin{tabular}{llccc}
\toprule
Model & Condition & $\kappa$ & Emb $\Delta\kappa$ & Gran $\Delta\kappa$ \\
\midrule
GPT-5.2 & \textsc{RAG\_no\_finetune} & 0.479 & & \\
GPT-5.2 & \textsc{RAG\_finetuned\_chunk} & 0.454 & $-0.025$ & \\
GPT-5.2 & \textsc{RAG\_finetuned\_utt} & \textbf{0.539} & & $+0.085$ \\
\midrule
Sonnet 4.6 & \textsc{RAG\_no\_finetune} & 0.498 & & \\
Sonnet 4.6 & \textsc{RAG\_finetuned\_chunk} & 0.501 & $+0.003$ & \\
Sonnet 4.6 & \textsc{RAG\_finetuned\_utt} & \textbf{0.580} & & $+0.079$ \\
\midrule
Qwen3-32b & \textsc{RAG\_no\_finetune} & 0.442 & & \\
Qwen3-32b & \textsc{RAG\_finetuned\_chunk} & 0.428 & $-0.014$ & \\
Qwen3-32b & \textsc{RAG\_finetuned\_utt} & \textbf{0.526} & & $+0.098$ \\
\bottomrule
\end{tabular}
\caption{Confound isolation on TalkMoves test ($k = 3$). Emb $\Delta\kappa$ is the effect of switching embeddings (Cohere $\to$ FT-BGE) at chunk-level granularity. Gran $\Delta\kappa$ is the effect of switching granularity (chunk $\to$ utterance) with FT-BGE.}
\label{tab:confound_tm}
\end{table}
\begin{table}[h]
\centering
\small
\begin{tabular}{llccc}
\toprule
Model & Condition & $\kappa$ & Emb $\Delta\kappa$ & Gran $\Delta\kappa$ \\
\midrule
GPT-5.2 & \textsc{RAG\_no\_finetune} & 0.505 & & \\
GPT-5.2 & \textsc{RAG\_finetuned\_chunk} & 0.530 & $+0.025$ & \\
GPT-5.2 & \textsc{RAG\_finetuned\_utt} & \textbf{0.659} & & $+0.129$ \\
\midrule
Sonnet 4.6 & \textsc{RAG\_no\_finetune} & 0.632 & & \\
Sonnet 4.6 & \textsc{RAG\_finetuned\_chunk} & 0.632$^\dagger$ & $+0.000$ & \\
Sonnet 4.6 & \textsc{RAG\_finetuned\_utt} & \textbf{0.743} & & $+0.111$ \\
\midrule
Qwen3-32b & \textsc{RAG\_no\_finetune} & 0.590 & & \\
Qwen3-32b & \textsc{RAG\_finetuned\_chunk} & 0.545 & $-0.045$ & \\
Qwen3-32b & \textsc{RAG\_finetuned\_utt} & \textbf{0.694} & & $+0.149$ \\
\bottomrule
\end{tabular}
\caption{Confound isolation on Eedi test ($k = 3$). $^\dagger$Sonnet \textsc{RAG\_finetuned\_chunk} is based on an incomplete run ($\sim$253 records short). Embedding model effect (Emb $\Delta\kappa$) ranges from $-0.045$ to $+0.025$. Granularity effect (Gran $\Delta\kappa$) ranges from $+0.079$ to $+0.149$.}
\label{tab:confound_eedi}
\end{table}
Across all six model-dataset combinations, switching the embedding model from Cohere to fine-tuned BGE while retaining chunk-level indexing yields negligible or slightly negative $\Delta\kappa$ (range: $-0.045$ to $+0.025$). In contrast, switching from chunk-level to utterance-level indexing with the same fine-tuned BGE model consistently produces large gains (range: $+0.079$ to $+0.149$). This confirms that utterance-level indexing is the primary driver of annotation improvement across all LLM backbones. Chunk-level embeddings average over multiple utterances that may serve different pedagogical functions, diluting the label-specific signal needed for accurate retrieval. Utterance-level indexing preserves this signal while still returning the full parent chunk as context for the LLM.
\end{document}